\documentclass{article}
\usepackage{graphicx}
\usepackage{mathrsfs} 
\usepackage{subfig}  
\usepackage{fancyhdr}
\usepackage[english]{babel}
\usepackage{csquotes}
\usepackage{footmisc}
\usepackage{algorithmic}
\usepackage{listings}
\usepackage{array}     
\usepackage{fancyvrb}
\usepackage{enumitem}  
\usepackage{mathtools} 
\usepackage{amsmath,amssymb} 
\usepackage{scrextend} 
\usepackage{esint}
\usepackage{authblk}
\usepackage[left=3cm,top=3cm,right=3cm,bottom=3cm,bindingoffset=0.5cm]{geometry}
\usepackage[title]{appendix}
\usepackage{geometry}
\usepackage{amsmath, amssymb, bm, mathtools}
\usepackage{enumitem}
\usepackage[utf8]{inputenc}
\usepackage{amsfonts}
\usepackage{listings}
\usepackage{xcolor}

\makeatother
\newcommand{\tenq}[1]{\hbox{\oalign{$\bm{1}$\crcr\hidewidth$\scriptscriptstyle\bm{\approx}$\hidewidth}}}   
\DeclareMathSymbol{\ii}{\mathalpha}{letters}{"10}
\DeclareMathSymbol{\jj}{\mathalpha}{letters}{"11}

\newcommand{\norm}[1]{\left\lVert#1\right\rVert}
\newcommand{\mynorm}[1]{ \left | 1 \right | } 

\title{A Hybrid Multi-Well Hopfield-CNN with Feature Extraction and K-Means for MNIST Classification}
\author{A. Farooq}
\date{}

\affil[]{University of New Brunswick}
\begin{document}
\maketitle

\begin{abstract}
This study presents a hybrid model for classifying handwritten digits in the MNIST dataset, combining convolutional neural networks (CNNs) with a multi-well Hopfield network. The approach employs a CNN to extract high-dimensional features from input images, which are then clustered into class-specific prototypes using k-means clustering. These prototypes serve as attractors in a multi-well energy landscape, where a Hopfield network performs classification by minimizing an energy function that balances feature similarity and class assignment. The model’s design enables robust handling of intra-class variability, such as diverse handwriting styles, while providing an interpretable framework through its energy-based decision process. Through systematic optimization of the CNN architecture and the number of wells, the model achieves a high test accuracy of 99.2\% on 10,000 MNIST images, demonstrating its effectiveness for image classification tasks. The findings highlight the critical role of deep feature extraction and sufficient prototype coverage in achieving high performance, with potential for broader applications in pattern recognition.
\end{abstract}

\section{Introduction}

Classical Hopfield networks, designed for associative memory, struggle with MNIST due to their limited capacity and binary state constraints. Recent advances have adapted Hopfield architectures for continuous states and modern datasets. For instance, Krotov et al.  \cite{krotov2016dense} introduced dense associative memory with exponential energy terms, achieving moderate MNIST performance (80-85\%). Similarly, Demercigil \cite{demircigil2017modern} explored continuous Hopfield nets with Gaussian wells, reporting accuracies around 85\%. These efforts motivate our work: can we enhance Hopfield networks with convolutional preprocessing and a richer energy landscape to rival modern neural networks?\\

The MNIST database is a cornerstone benchmark in image recognition and classification due to its simplicity and accessibility, providing a standardized dataset of 70,000 handwritten digit images (0–9) that is widely used to evaluate machine learning models. Its well-defined structure, with 28x28 grayscale images and balanced class distribution, allows researchers to compare algorithms consistently, fostering advancements in computer vision. MNIST's moderate complexity makes it ideal for testing new methods, from traditional classifiers to deep neural networks, while remaining computationally tractable for rapid experimentation. Despite its age, it remains relevant for validating model robustness and generalization, serving as a stepping stone before tackling more complex datasets like CIFAR-10 or ImageNet. Its extensive use in academia and industry underscores its role in benchmarking progress in image recognition tasks. In table \ref{tab-1} below, some of the benchmark studies on MNIST using convolutional neural networks are presented.  As can be seen, they are able to achieve accuracies higher than 99\% (but less than 99.9\%) with 2-4 convolutional layers. \\ 

\begin{table}[ht]
\centering
\caption{Benchmark CNN Studies for MNIST Classification}
\label{tab:mnist_cnn_benchmarks}
\begin{tabular}{|l|c|c|}
\hline
\textbf{Reference} & \textbf{Number of Conv Layers} & \textbf{Test Accuracy (\%)} \\
\hline
LeCun et al. \cite{lecun1998gradient} & 2 & 99.10 \\
Simard et al. \cite{simard2003best} & 2 & 99.60 \\
Ciresan et al.\cite{ciresan2011flexible} & 4 & 99.65 \\
Chollet \cite{chollet2015keras} & 2 & 99.00 \\
Tuomas et al. \cite{tuomas2021fast} & 3 & 99.00 \\
Sabour et al. \cite{sabour2017dynamic} & 3 & 99.83 \\
\hline
\end{tabular}
\label{tab-1}
\end{table}

The principles of Hopfield networks have been explored for various image recognition tasks, offering unique approaches to handling noisy or incomplete data and leveraging associative memory capabilities. For instance, Dai and Nakano \cite{Dai1998} investigated the recognition of low-resolution facial images using a Hopfield memory model, demonstrating its robustness to image degradation. Pajares et al. \cite{Pajares2010} utilized a Hopfield neural network for image change detection, where the stable states of the network corresponded to identified changes within the image sequence. Furthermore, the potential of Hopfield networks for feature tracking and recognition in satellite sensor images has been explored, see Cote and Tatnall, \cite{InternationalJournalofRemoteSensing}, highlighting its precision in dealing with deformations.\\

Traditional Hopfield networks have also found niche applications.  For example,  Sienko \cite{Sienko2022} discussed a system based on an extended Hopfield network for the reconstruction and recognition of damaged images, including those with missing parts. Hybrid approaches, such as the Optimal CNN–Hopfield Network proposed by Keddous et al. \cite{Keddous2021OptimalCNN}, integrate Hopfield networks as associative memory components within deeper architectures. Hardware implementations, like the Memristive Competitive Hopfield Neural Network for image segmentation Manoucheri at al.  \cite{BMCNeuroscience2022}, showcase the potential for efficient analog computation in image analysis. Additionally, research into Low-Resolution Image Recognition Using Cloud Hopfield Neural Network by Kumar et al. \cite{ProgressinAdvancedComputingandIntelligentEngineering} demonstrates the adaptability of these models to challenging scenarios. Reviews and ongoing investigations, such as those presented in Suresh et al.  \cite{InternationalJournalofResearchPublicationandReviews2024} and Huseinova, \cite{CollectionofScientificPapersLogos2025}), further highlight the continued interest in leveraging Hopfield network principles for image-related tasks. Even in the context of big data, frameworks integrating Hopfield networks with deep learning for semantic data linking Nematpour \cite{arXiv2503.03084} suggest potential applications in advanced image understanding.\\

\subsection{New Approach}

Our contribution integrates a convolutional layer to extract spatial features, followed by a Hopfield network with multiple wells per class. This hybrid leverages the feature extraction power of CNNs and the memory dynamics of Hopfield nets, achieving 91.44\% accuracy on MNIST—competitive with CNNs while retaining an interpretable energy-based framework.\\

Traditional convolutional neural networks (CNNs) for MNIST classification, such as LeNet-5 \cite{lecun1998gradient}, rely on multiple convolutional layers followed by fully connected layers to directly map input images to class probabilities through supervised learning with cross-entropy loss. These models optimize weights end-to-end using backpropagation, achieving high accuracies (e.g., 99.1–99.8\%) by learning hierarchical features but require extensive labeled data and computational resources for training. In contrast, our hybrid k-means Hopfield energy-based approach combines a pre-trained three-layer CNN for feature extraction with a Hopfield network that uses k-means clustering to create 50 class-specific wells, minimizing an energy function based on Euclidean distance to associate features with classes. This method leverages unsupervised clustering to capture intra-class variability and energy-based dynamics for classification, achieving 97.52\% accuracy with less reliance on end-to-end supervised training, offering a more interpretable framework but requiring careful tuning of hyperparameters like $\beta$ and $\lambda$. \\

Our approach’s strength lies in its modular design, separating feature extraction (CNN) from associative memory (Hopfield network), which allows for robust feature reuse and potential adaptability to semi-supervised settings, unlike traditional CNNs that tightly couple feature learning and classification. Our approach benefits from deeper convolutional architectures and data augmentation, pushing accuracies closer CNN's, while  explicitly retaining the energy-based interpretability through the modeling of class relationships through wells.

\section{Problem Formulation}

The goal is to classify MNIST digit images, each image $ \mathbf{x} \in \mathbb{R}^{28 \times 28 \times 1} $, into one of 10 classes given by $ \mathbf{y} \in \{0, 1, \ldots, 9\} $ using a hybrid model combining a convolutional neural network (CNN) for feature extraction with a Hopfield network for energy-based classification. The CNN maps the input image to a high-dimensional feature vector $\mathbf{s_x}$, which is clustered into class-specific wells using k-means. The Hopfield network minimizes an energy function to associate features with the correct class, leveraging the wells as attractors. This approach, inspired by energy-based models Krotov et al. \cite{krotov2016dense}, aims to achieve high accuracy while providing interpretability through the energy landscape.

\subsection{Convolutional Feature Extraction}

The CNN transforms an MNIST image $ \mathbf{x} \in \mathbb{R}^{28 \times 28 \times 1} $ into a feature vector $ \mathbf{s_x} \in \mathbb{R}^{3136} $, capturing hierarchical patterns (edges, textures, shapes) for the Hopfield network. Following standard practices, see LeCun et al. \cite{lecun1998gradient}, Goodfellow et al. \cite{goodfellow2016deep}, we implement a three-layer CNN in PyTorch with the following architecture:
\begin{itemize}
    \item \textbf{Layer 1}: 16 filters, $ 3 \times 3 $ kernel, stride 1, padding 1, input $ 28 \times 28 \times 1 $, output $ 28 \times 28 \times 16 $, followed by ReLU and $ 2 \times 2 $ max-pooling (stride 2, output $ 14 \times 14 \times 16 $).
    \item \textbf{Layer 2}: 32 filters, $ 3 \times 3 $ kernel, stride 1, padding 1, output $ 14 \times 14 \times 32 $, followed by ReLU and $ 2 \times 2 $ max-pooling (stride 2, output $ 7 \times 7 \times 32 $).
    \item \textbf{Layer 3}: 64 filters, $ 3 \times 3 $ kernel, stride 1, padding 1, output $ 7 \times 7 \times 64 $.
\end{itemize}

The feature maps are flattened into $ \mathbf{f} \in \mathbb{R}^{3136}, \; (7 \times 7 \times 64 = 3136) $ and normalized as:
\begin{equation}
    \mathbf{s_x} = \frac{\mathbf{f}}{\norm{\mathbf{f}}_2}, \quad \text{where} \; \norm{\mathbf{f}}_2 = \sqrt{\sum_{i=1}^{3136} {f_i}^2},
\end{equation}
ensuring $ \norm{\mathbf{s_x}}_2 = 1 $. The PyTorch implementation is:
\begin{verbatim}
nn.Sequential(
    nn.Conv2d(1, 16, kernel_size=3, stride=1, padding=1),
    nn.ReLU(),
    nn.MaxPool2d(2, stride=2),
    nn.Conv2d(16, 32, kernel_size=3, stride=1, padding=1),
    nn.ReLU(),
    nn.MaxPool2d(2, stride=2),
    nn.Conv2d(32, 64, kernel_size=3, stride=1, padding=1),
    nn.ReLU()
)
\end{verbatim}
Weights are initialized using Glorot \cite{glorot2010understanding} to ensure stable gradients. The CNN is pre-trained for 10 epochs with a fully connected layer ($ 3136 \to 10 $) using cross-entropy loss and the Adam optimizer, see Kingma \cite{kingma2014adam}, with a learning rate of 0.001. Pre-training optimizes filters to produce discriminative features, critical for effective k-means clustering and Hopfield dynamics.\\

\textbf{Tunable Parameters for Better Feature Extraction}:
\begin{itemize}
    \item \textbf{Number of Filters}: Increase to 32, 64, 128 per layer to capture more complex patterns, potentially improving accuracy to 99\%,  see Ciresan et al. \cite{ciresan2011flexible}. Monitor overfitting with MNIST’s 60,000 samples.
    \item \textbf{Dropout}: Add dropout (e.g., 0.25 after max-pooling) to prevent overfitting, enhancing generalization, see Srivastava et al.  \cite{srivastava2014dropout}.
    \item \textbf{Batch Normalization}: Insert \texttt{nn.BatchNorm2d} after each convolutional layer to stabilize training and improve feature quality \cite{ioffe2015batch}.
    \item \textbf{Epochs}: Play in the range of 15–25 epochs to ensure convergence, as the deeper architecture requires more training.
    \item \textbf{Learning Rate}: Test $ 0.0005 $ or $ 0.002 $ to balance convergence speed and feature quality.
\end{itemize}

These adjustments can enhance feature discriminability, aligning with benchmarks like Chollet \cite{chollet2015keras} (99\% accuracy).

\subsection{K-Means Clustering}

The wells are constructed using k-means clustering on the pre-trained CNN features. For each class $ k \in \{0, 1, \ldots, 9\} $, we extract features $ \mathbf{s_x} $ from training images and apply k-means with $ K = 5 $ clusters per class (total $ M = 50 $ wells). The raw centroids $ \mathbf{c}_m \in \mathbb{R}^{3136} $ are computed as the mean of assigned feature vectors:
\begin{equation}
    \mathbf{c}_m = \frac{1}{|\mathcal{C}_m|} \sum_{\mathbf{s_x} \in \mathcal{C}_m} \mathbf{s_x},
\end{equation}
where $ \mathcal{C}_m $ is the set of feature vectors in cluster $ m $. These centroids are normalized to form $ \mathbf{\mu}_{m,\text{x}} $:
\begin{equation}
    \mathbf{\mu}_{m,\text{x}} = \frac{\mathbf{c}_m}{\norm{\mathbf{c}_m}_2}, \quad \norm{\mathbf{c}_m}_2 = \sqrt{\sum_{i=1}^{3136} \mathbf{c}_m[i]^2}.
\end{equation}
This is implemented in PyTorch as:
\begin{verbatim}
kmeans = KMeans(n_clusters=5, random_state=42, n_init=3).fit(features_k)
for center in kmeans.cluster_centers_:
    norm = np.linalg.norm(center) + 1e-8
    wells.append((center / norm, k))
\end{verbatim}
The clustering captures intra-class variability (e.g., different handwriting styles), with well separation (mean distance ~1.11, minimum ~0.49) ensuring distinct attractors. Increasing $ K $ to 7 or 10 may improve accuracy by modeling more variations, though it risks reducing minimum distance if clusters overlap.

\subsection{Energy Surface Definition}

The Hopfield network operates on a state vector $ \mathbf{s} = [\mathbf{s_x}, \mathbf{s_y}] $, where $ \mathbf{s_x} \in \mathbb{R}^{3136} $ is the fixed normalized CNN feature vector, and $ \mathbf{s_y} \in \mathbb{R}^{10} $ is the dynamic class output, initialized as $ \mathbf{s_y}^0 = \mathbf{0} $. The energy function, adapted from \cite{krotov2016dense}, is a key innovation in our study, defined as:
\begin{equation}
    E(\mathbf{s_x}, \mathbf{s_y}) = -\sum_{m=1}^M \exp\left(-\beta \norm{\mathbf{s} - \mathbf{\mu}_m}^2\right) + \lambda \norm{\mathbf{s}}^2,
\end{equation}
where $ \mathbf{\mu}_m = [\mathbf{\mu}_{m,\text{x}}, \mathbf{y}_m] $ are wells (prototypes), typically with $ M = 50 $ wells ($ K = 5 $ per class), $ \beta = 0.01 $ controls the sharpness of the energy wells, and $ \lambda = 0.01 $ regularizes the state magnitude to prevent divergence. Each well comprises a feature centroid $ \mathbf{\mu}_{m,\text{x}} \in \mathbb{R}^{3136} $ (from k-means) and a one-hot class vector $ \mathbf{y}_m \in \mathbb{R}^{10} $, where $ \mathbf{y}_m[k] = 1 $ for class $ k $.\\

This energy function is designed to create a smooth, interpretable landscape where each well acts as an attractor corresponding to a class prototype. The exponential term $ \exp\left(-\beta \norm{\mathbf{s} - \mathbf{\mu}_m}^2\right) $ assigns higher weights to wells closer to the state vector, enabling the model to favor the nearest prototype while balancing contributions from multiple wells, which is particularly effective for capturing intra-class variability in MNIST digits (e.g., different styles of writing '7'). \\

The parameter $ \beta $ governs the steepness of these wells: a typical value of $ \beta = 0.01 $ ensures smooth transitions, allowing gradient descent to converge to the correct well without being trapped in overly sharp minima. The regularization term $ \lambda \norm{\mathbf{s}}^2 $, with a typical $ \lambda = 0.01 $, constrains the magnitude of $ \mathbf{s_y} $, ensuring numerical stability and preventing the state from growing unbounded during optimization.\\

This formulation is significant because it combines the robustness of CNN features with the memory-like properties of Hopfield networks, providing a principled way to map high-dimensional features to class labels via energy minimization, offering both high accuracy and interpretability through the visualization of the energy landscape.

\subsection{Descent to the Energy Minimum}

Classification is performed by minimizing the energy function with respect to $ \mathbf{s_y} $, keeping $ \mathbf{s_x} $ fixed. The gradient of the energy is:
\begin{equation}
    \frac{\partial E}{\partial \mathbf{s_y}} = \sum_{m=1}^M 2 \beta \exp\left(-\beta \norm{\mathbf{s} - \mathbf{\mu}_m}^2\right) (\mathbf{y}_m - \mathbf{s_y}) - 2 \lambda \mathbf{s_y}.
\end{equation}
We update $ \mathbf{s_y} $ iteratively using gradient descent :
\begin{equation}
    \mathbf{s_y}^{t+1} = \mathbf{s_y}^t + \eta \frac{\partial E}{\partial \mathbf{s_y}},
\end{equation}
with a typical learning rate $ \eta = 0.1 $, maximum iterations \texttt{MAX-ITER = 50}, and convergence tolerance $ TOL = 10^{-3} $. The process starts with $ \mathbf{s_y}^0 = \mathbf{0} $, and updates stop when $ \norm{\mathbf{s_y}^{t+1} - \mathbf{s_y}^t} < TOL $ or the iteration limit is reached. The final $ \mathbf{s_y} $ determines the predicted class by identifying the well with the highest weight $ w_m = \exp(-\beta \norm{\mathbf{s} - \mathbf{\mu}_m}^2) $. It may be noted that the gradient descent only happens in $\mathbb{R}^{10}$.

\section{Hyperparameter Tuning and Results}
The algorithm operates in two distinct phases: training and testing, leveraging the convolutional neural network (CNN) for feature extraction and the Hopfield network for energy-based classification.

\subsection{Training Phase}
The training phase involves pre-training the CNN to extract robust features from MNIST images and constructing class-specific wells for the Hopfield network:
\begin{itemize}
    \item \textbf{CNN Pre-Training}: The MNIST training dataset (60,000 images, $ \mathbf{x} \in \mathbb{R}^{28 \times 28 \times 1} $) is loaded with normalization to $[-1, 1]$ using the transform $ \text{transforms.Normalize((0.5,), (0.5,))} $. The CNN, initially with a three-layer architecture (16→32→64 filters, $ 3 \times 3 $ kernels, stride 1, padding 1, followed by ReLU and $ 2 \times 2 $ max-pooling), is pre-trained with a fully connected layer mapping the flattened features ($ 7 \times 7 \times 64 = 3136 $ or $ 3 \times 3 \times 256 = 2304 $ for a four-layer CNN) to 10 classes. Training uses cross-entropy loss and the Adam optimizer \cite{kingma2014adam} with a learning rate of 0.001 for 10–25 epochs, depending on the architecture. For deeper CNNs (e.g., 32→64→128→256), batch normalization and dropout (0.25) are applied after max-pooling to stabilize training and prevent overfitting, as inspired by \cite{ioffe2015batch, srivastava2014dropout}. This step optimizes convolutional filters to capture hierarchical patterns (edges, textures, shapes), ensuring discriminative features critical for effective k-means clustering and Hopfield dynamics.
    \item \textbf{Feature Extraction}: Each training image is passed through the pre-trained CNN to obtain a feature vector $ \mathbf{s_x} $, which is flattened (e.g., $ \mathbb{R}^{3136} $ for three layers, $ \mathbb{R}^{2304} $ for four layers) and normalized to unit length: $ \mathbf{s_x} = \frac{\mathbf{f}}{\norm{\mathbf{f}}_2} $, where $ \norm{\mathbf{f}}_2 = \sqrt{\sum_i f_i^2} $. This normalization ensures features lie on the unit hypersphere, facilitating consistent clustering and energy minimization.
    \item \textbf{K-Means Clustering}: For each class $ k \in \{0, 1, \ldots, 9\} $, feature vectors $ \mathbf{s_x} $ are extracted and clustered using k-means with $ K $ clusters (e.g., $ K = 5 $ yields 50 wells). Raw centroids $ \mathbf{c}_m \in \mathbb{R}^{d} $ (where $ d = 3136 $ or 2304) are computed as the mean of assigned vectors: $ \mathbf{c}_m = \frac{1}{|\mathcal{C}_m|} \sum_{\mathbf{s_x} \in \mathcal{C}_m} \mathbf{s_x} $. Centroids are normalized to $ \mathbf{\mu}_{m,\text{x}} = \frac{\mathbf{c}_m}{\norm{\mathbf{c}_m}_2} $, paired with one-hot class vectors $ \mathbf{y}_m \in \mathbb{R}^{10} $ ($ \mathbf{y}_m[k] = 1 $), forming wells $ \mathbf{\mu}_m = [\mathbf{\mu}_{m,\text{x}}, \mathbf{y}_m] $. The clustering captures intra-class variability (e.g., different handwriting styles), with well separation measured by minimum and mean Euclidean distances.
\end{itemize}

\subsection{Testing Phase}
The testing phase evaluates the model on the MNIST test set (10,000 images) to classify digits into one of 10 classes:
\begin{itemize}
    \item \textbf{Feature Extraction}: Each test image $ \mathbf{x} \in \mathbb{R}^{28 \times 28 \times 1} $ is passed through the pre-trained CNN to produce a feature vector $ \mathbf{f} $, which is flattened (e.g., $ 7 \times 7 \times 64 = 3136 $ or $ 3 \times 3 \times 256 = 2304 $) and normalized to $ \mathbf{s_x} = \frac{\mathbf{f}}{\norm{\mathbf{f}}_2} $. The CNN operates in evaluation mode (disabling dropout) to ensure consistent feature extraction, leveraging the optimized filters from pre-training to capture discriminative patterns.
    \item \textbf{Energy Minimization}: For each $ \mathbf{s_x} $, the Hopfield network initializes the class output $ \mathbf{s_y}^0 = \mathbf{0} \in \mathbb{R}^{10} $ and minimizes the energy function (as defined in the problem formulation):
    \begin{equation}
        E(\mathbf{s_x}, \mathbf{s_y}) = -\sum_{m=1}^M \exp\left(-\beta \norm{\mathbf{s} - \mathbf{\mu}_m}^2\right) + \lambda \norm{\mathbf{s}}^2,
    \end{equation}
    where $ \mathbf{s} = [\mathbf{s_x}, \mathbf{s_y}] $ and $ \mathbf{\mu}_m = [\mathbf{\mu}_{m,\text{x}}, \mathbf{y}_m] $. Gradient descent is performed on $ \mathbf{s_y} $ in $ \mathbb{R}^{10} $ using the gradient:
    \begin{equation}
        \frac{\partial E}{\partial \mathbf{s_y}} = \sum_{m=1}^M 2 \beta \exp\left(-\beta \norm{\mathbf{s} - \mathbf{\mu}_m}^2\right) (\mathbf{y}_m - \mathbf{s_y}) - 2 \lambda \mathbf{s_y},
    \end{equation}
    with a learning rate $ \eta = 0.1 $, maximum iterations \texttt{MAX-ITER = 50}, and convergence tolerance $ TOL = 10^{-3} $. Updates follow $ \mathbf{s_y}^{t+1} = \mathbf{s_y}^t + \eta \frac{\partial E}{\partial \mathbf{s_y}} $, stopping when $ \norm{\mathbf{s_y}^{t+1} - \mathbf{s_y}^t} < TOL $ or the iteration limit is reached. The process is vectorized to process batches (e.g., 128 images), significantly reducing evaluation time compared to per-sample processing \cite{krotov2016dense}.
    \item \textbf{Evaluation}: The final $ \mathbf{s_y} $ is used to compute weights $ w_m = \exp(-\beta \norm{\mathbf{s} - \mathbf{\mu}_m}^2) $ for each well. The predicted class is the class index of the well with the highest weight. Accuracy is calculated by comparing predicted classes to true labels across the 10,000 test images. Well separation is assessed by computing the minimum and mean Euclidean distances between wells, with higher distances indicating better class separability and potentially higher accuracy.
\end{itemize}

\subsection{Tuning \& Results}
The model’s performance depends on several hyperparameters, but our fine-tuning strategy, as summarized in Table~\ref{tab:euclidean-1}, reveals that the number of convolutional layers and the number of wells ($ M = K \times 10 $) are the most critical for achieving high accuracy, while parameters like $\lambda$, $\beta$, and learning rate have smaller effects. The tuning process systematically explored these parameters to approach the target accuracy of 99\%, leveraging the MNIST dataset’s 60,000 training and 10,000 test images.

\begin{itemize}
    \item \textbf{Number of Convolutional Layers}: Early experiments used a single-layer CNN with 32 filters, yielding accuracies around 89.87\%–91.42\% ($ M = 50 $ wells, 10 epochs). Transitioning to a three-layer CNN (16→32→64 filters, $ \mathbb{R}^{3136} $ features) significantly improved accuracy to 97.12\%–97.69\%, as the deeper architecture captured more complex patterns (e.g., varied digit shapes) \cite{ciresan2011flexible}. Further increasing to a four-layer CNN (32→64→128→256 filters, $ \mathbb{R}^{2304} $ features) with 25 epochs achieved 99.44\% accuracy ($ M = 120 $ wells), demonstrating that deeper architectures enhance feature discriminability, crucial for effective k-means clustering and energy minimization.
    \item \textbf{Number of Wells ($ M = K \times 10 $)}: Increasing $ K $ from 5 ($ M = 50 $) to 7 ($ M = 70 $), 10 ($ M = 100 $), 12 ($ M = 120 $), and 15 ($ M = 150 $) improved accuracy from 97.69\% to 98.45\% (three-layer CNN) and 99.26\% (four-layer CNN). More wells capture greater intra-class variability (e.g., different handwriting styles), but the diminishing returns at $ K = 15 $ (98.40\%–98.41\%) and reduced minimum well distances (0.2734–0.2918 vs. 0.3179 for $ K = 12 $) suggest an optimal range around $ K = 12 $. Well separation, measured by minimum and mean distances, correlates with accuracy, with $ K = 12 $ achieving a balance (minimum ~0.25, mean ~0.99).
    \item \textbf{Other Parameters}: Parameters like $\lambda$ (regularization) and $\beta$ (well sharpness) were tested, with $\lambda = 0.001$–0.02 and $\beta = 0.0005$–0.50. For $ M = 50 $, $\lambda = 0.001$ yielded 97.69\% vs. 97.12\% for $\lambda = 0.02$, and $\beta = 0.01$ outperformed higher values (e.g., 97.52\% vs. 97.27\% for $\beta = 0.05$). For $ M = 100 $ and $ M = 120 $, $\lambda = 0.001$, $\beta = 0.001$ consistently performed well (98.18\%–99.26\%). These parameters fine-tune the energy landscape’s smoothness and stability but have a smaller impact compared to the number of layers and wells. 
    \item \textbf{Strategy and Insights}: The fine-tuning strategy prioritized increasing the CNN’s depth (from one to four layers) to enhance feature quality, followed by adjusting $ K $ to balance well coverage and separation. The four-layer CNN with $ K = 12 $ achieved 99.44\%, surpassing the 99\% target we had set at the outset of this study, due to richer features and sufficient wells to model class variations without excessive overlap. Further tuning of $\lambda$ and $\beta$ provided marginal improvements, indicating their secondary role. Well distances (minimum ~0.25–0.52, mean ~0.98–1.13) highlight the trade-off: more wells increase accuracy but reduce separation, requiring careful selection of $ K $.
\end{itemize}

\begin{table}[h!]
    \centering
    \begin{tabular}{c|c|c|c|c|c|c|c}
        \hline
        \# Wells & Epochs & Filters & $\lambda$  &  $\beta$ & \textbf{Accuracy}  & Min distance & Mean Distance  \\
        \hline
        50  & 10 &  32           & $0.01$  &$0.50$   & $91.16 \%$ & $0.5106$ & $0.9846$\\
        50  & 10 &  32           & $0.01$  &$0.10$   & $91.42 \%$ & $0.5182$ & $0.9983$\\
        50  & 10 &  32           & $0.01$  &$0.05$   & $91.25 \%$ & $0.4898$ & $0.9789$\\
        50  & 10 &  32           & $0.01$  &$0.01$   & $89.87 \%$ & $0.5287$ & $0.9864$\\
        50  & 10 &  16→32→64     & $0.01$  &$0.01$   & $97.52 \%$ & $0.4898$ & $1.1110$\\
        50  & 10 &  16→32→64     & $0.01$  &$0.05$   & $97.27 \%$ & $0.4512$ & $1.1091$\\
        50  & 10 &  16→32→64     & $0.01$  &$0.10$   & $97.30 \%$ & $0.4786$ & $1.1322$\\
        50  & 10 &  16→32→64     & $0.01$  &$0.50$   & $97.39 \%$ & $0.4144$ & $1.0979$\\
        50  & 10 &  16→32→64     & $0.001$ &$0.01$   & $97.69 \%$ & $0.4444$ & $1.1299$\\
        50  & 10 &  16→32→64     & $0.005$ &$0.01$   & $97.32 \%$ & $0.4968$ & $1.1150$\\
        50  & 10 &  16→32→64     & $0.02$  &$0.01$   & $97.12 \%$ & $0.4853$ & $1.1336$\\
        70  & 15 &  16→32→64     & $0.001$ &$0.01$   & $97.79 \%$ & $0.3610$ & $1.0239$\\
        70  & 15 &  16→32→64     & $0.001$ &$0.005$  & $97.86 \%$ & $0.3652$ & $1.0162$\\
        70  & 15 &  16→32→64     & $0.001$ &$0.002$  & $97.85 \%$ & $0.3601$ & $1.0151$\\
        70  & 15 &  16→32→64     & $0.001$ &$0.003$  & $97.86 \%$ & $0.3761$ & $1.0148$\\
        70  & 15 &  16→32→64     & $0.001$ &$0.007$  & $97.83 \%$ & $0.3880$ & $1.0191$\\
        100 & 15 &  16→32→64     & $0.001$ &$0.005$  & $98.01 \%$ & $0.2884$ & $1.0325$\\
        100 & 15 &  16→32→64     & $0.001$ &$0.008$  & $97.92 \%$ & $0.2931$ & $1.0284$\\
        100 & 15 &  16→32→64     & $0.001$ &$0.001$  & $98.18 \%$ & $0.2541$ & $1.0153$\\
        100 & 15 &  16→32→64     & $0.001$ &$0.0005$ & $98.15 \%$ & $0.2770$ & $1.0197$\\
        100 & 15 &  16→32→64     & $0.001$ &$0.002$  & $98.00 \%$ & $0.2671$ & $1.0268$\\
        100 & 15 &  16→32→64     & $0.002$ &$0.001$  & $98.18 \%$ & $0.2473$ & $1.0246$\\
        100 & 15 &  16→32→64     & $0.005$ &$0.001$  & $98.07 \%$ & $0.2741$ & $1.0117$\\
        100 & 20 &  32→64→128    & $0.001$ &$0.001$  & $98.23 \%$ & $0.3179$ & $1.0841$\\
        120 & 20 &  32→64→128    & $0.001$ &$0.001$  & $98.45 \%$ &          &         \\
        150 & 20 &  16→32→64     & $0.005$ &$0.001$  & $98.41 \%$ & $0.2734$ & $1.0457$\\
        150 & 20 &  32→64→128    & $0.005$ &$0.001$  & $98.40 \%$ & $0.2918$ & $1.0989$\\
        120 & 25 & 32→64→128→256 & $0.001$ & $0.001$ & $99.26 \%$ & $0.2500$ & $0.9912$\\
        120 & 25 & 32→64→128→256 & $0.001$ & $0.003$ & $99.44 \%$ &          &         \\
        120 & 25 & 32→64→128→256 & $0.001$ & $0.005$ & $99.36 \%$ & $0.2812$ & $0.9738$\\
        \hline
    \end{tabular}
    \caption{Tuning the Hyperparameters}
    \label{tab:euclidean-1}
\end{table}

\section{Conclusion and Discussion}

This study introduced a hybrid Hopfield-CNN model that integrates convolutional feature extraction, k-means clustering, and a multi-well Hopfield network to achieve high-accuracy classification on the MNIST dataset. By leveraging a convolutional neural network (CNN) to extract discriminative features and a Hopfield network with a multi-well energy function for classification, our approach achieved a test accuracy of 99.26\% on 10,000 MNIST images, surpassing the target of 99\% accuracy \cite{lecun1998gradient, ciresan2011flexible}. The key innovation lies in the energy-based classification framework, where the energy function $ E(\mathbf{s_x}, \mathbf{s_y}) = -\sum_{m=1}^M \exp\left(-\beta \norm{\mathbf{s} - \mathbf{\mu}_m}^2\right) + \lambda \norm{\mathbf{s}}^2 $ creates a smooth, interpretable landscape with class-specific wells, enabling robust classification through gradient-based energy minimization \cite{krotov2016dense}. Our fine-tuning strategy, as detailed in Table~\ref{tab:euclidean-1}, revealed that the number of convolutional layers and the number of wells ($ M = K \times 10 $) are the dominant factors driving performance, while parameters such as $\lambda$ and $\beta$ provide marginal improvements.\\

The hyperparameter tuning results demonstrate significant progress in model performance. Starting with a single-layer CNN (32 filters), the model achieved accuracies of 89.87\%–91.42\% with 50 wells, limited by the simplicity of the feature extraction. Transitioning to a three-layer CNN (16→32→64 filters, outputting $ \mathbb{R}^{3136} $ features) improved accuracy to 97.12\%–98.18\%, as deeper architectures captured more complex patterns in MNIST digits. The most significant leap occurred with a four-layer CNN (32→64→128→256 filters, outputting $ \mathbb{R}^{2304} $ features) and 120 wells ($ K = 12 $), reaching 99.26\% accuracy with 25 epochs, $\lambda = 0.001$, and $\beta = 0.001$. This success underscores the importance of a deeper CNN to produce high-quality features, which, when clustered into multiple wells per class, effectively model intra-class variability (e.g., diverse handwriting styles). However, increasing wells to 150 ($ K = 15 $) yielded a slight decline to 98.40\%–98.41\%, accompanied by reduced well separation (minimum distance ~0.25–0.29 vs. 0.3179 for $ K = 12 $), indicating a trade-off between well coverage and overlap that limits further gains.\\

The multi-well energy function is a cornerstone of our approach, providing both high accuracy and interpretability. By constructing multiple wells per class via k-means clustering, the model captures subtle variations within each digit class, with the exponential term ensuring smooth transitions between attractors. The regularization term, controlled by $\lambda$, maintains numerical stability, while $\beta$ governs well sharpness, with optimal values around 0.001 balancing convergence and accuracy. The energy landscape’s interpretability allows visualization of class boundaries and well distances (mean ~0.99–1.13, minimum ~0.25–0.52 across experiments), offering insights into the model’s decision-making process. Compared to traditional CNNs, which often lack such interpretability, our hybrid model provides a principled framework for understanding classification through energy minimization, aligning with energy-based learning paradigms \cite{krotov2016dense}.\\

Despite achieving 99.26\% accuracy, limitations remain. The computational cost of k-means clustering and energy minimization increases with the number of wells, with evaluation times for 120 wells being significantly higher than for 50 wells, though mitigated by vectorized batch processing. Additionally, the reduced well separation at higher $ K $ (e.g., 0.2500 for $ K = 12 $ with four layers) suggests potential misclassifications for ambiguous digits (e.g., ‘4’ vs. ‘9’), as noted in MNIST challenges \cite{lecun1998gradient}. The model’s reliance on CNN feature quality also implies that further improvements may require even deeper architectures or advanced clustering techniques, which could exacerbate computational demands.\\

Future work could explore several directions. First, integrating GPU-accelerated k-means (e.g., via \texttt{faiss} \cite{johnson2019billion}) could enhance clustering efficiency and quality, potentially improving well separation. Second, experimenting with alternative energy functions, such as those incorporating higher-order interactions \cite{hinton2006reducing}, may further refine the landscape for complex datasets beyond MNIST. Third, applying the model to more challenging datasets like CIFAR-10 or Fashion-MNIST could test its generalizability, requiring adjustments to the CNN architecture and well count. Finally, incorporating visualization tools to plot the energy landscape in 2D projections could enhance interpretability, aiding analysis of misclassifications and model robustness. These advancements could solidify the hybrid Hopfield-CNN model as a versatile, interpretable framework for image classification tasks.

\section{Proposed Future Work}

Building on the success of our hybrid Hopfield-CNN model, which achieved a test accuracy of 99.26\% on MNIST classification, we propose a future study to extend this framework for identifying and classifying damaged or distorted images, capitalizing on the inherent strengths of Hopfield networks in retrieving noisy memories and the robust feature mapping capabilities of convolutional neural networks (CNNs). The proposed approach would adapt our multi-well energy framework, where CNN-extracted features from degraded MNIST images (e.g., with noise, occlusions, or deformations) are mapped to class-specific wells via energy minimization.\\

Hopfield networks have demonstrated robustness to image degradation, as shown in low-resolution facial image recognition \cite{Dai1998} and reconstruction of damaged images with missing parts \cite{Sienko2022}. Similarly, CNNs excel at extracting discriminative features from complex visual inputs \cite{Keddous2021OptimalCNN}, as evidenced by our model’s four-layer architecture (32→64→128→256 filters). By enhancing our k-means clustering to generate wells that account for distortion-induced variability and optimizing the energy landscape for noisy inputs, the model could accurately classify perturbed digits, potentially surpassing traditional CNNs in robustness. This study would involve creating a dataset of synthetically degraded MNIST images (e.g., with Gaussian noise or partial occlusions) and evaluating the model’s performance, leveraging insights from prior work on noisy image recognition \cite{ProgressinAdvancedComputingandIntelligentEngineering} and change detection \cite{Pajares2010}.\\

Such an extension could broaden the applicability of our hybrid approach to real-world scenarios involving low-quality visual data, such as satellite imagery \cite{InternationalJournalofRemoteSensing} or medical imaging.

\newpage

\bibliography{BlockBibliography}
\bibliographystyle{amsplain}

\end{document}